\documentclass[10pt,twocolumn,letterpaper]{article}

\usepackage{iccv}
\usepackage{times}
\usepackage{epsfig}
\usepackage{graphicx}
\usepackage{amsmath}
\usepackage{amssymb}

\usepackage{tikz}
\usetikzlibrary{decorations.pathreplacing}
\usetikzlibrary{calc}
\usepackage{multirow}
\usepackage{cuted}
\usepackage{makecell}
\usepackage{booktabs}
\newcommand{\ra}[1]{\renewcommand{\arraystretch}{#1}}
\usepackage[utf8]{inputenc}

\DeclareMathOperator*{\median}{median}

\DeclareMathOperator*{\argmax}{argmax}

\usepackage[pagebackref=true,breaklinks=true,letterpaper=true,colorlinks,bookmarks=false]{hyperref}

\iccvfinalcopy 

\def\iccvPaperID{2239} 

\ificcvfinal\fi
\begin{document}

\title{Multi Target Tracking by Learning from Generalized Graph Differences}

\author{Håkan Ardö\\
Center for Mathematical Sciences\\
Lund University\\
{\tt\small ardo@maths.lth.se}
\and
Mikael Nilsson\\
Center for Mathematical Sciences\\
Lund University\\
{\tt\small micken@maths.lth.se}
}

\maketitle

\begin{abstract}
  Formulating the multi object tracking problem as a network flow optimization problem is a popular choice. In this paper an efficient way of learning the weights of such a network is presented. It separates the problem into one embedding of feasible solutions into a one dimensional feature space and one optimization problem. The embedding can be learned using standard SGD type optimization without relying on an additional optimizations within each step. Training data is produced by performing small perturbations of ground truth tracks and  representing them using generalized graph differences, which is an efficient way introduced to represent the difference between two graphs. The proposed method is evaluated on DukeMTMCT with competitive results.
\end{abstract}

\section{Introduction}

Detecting and tracking objects in video sequences is interesting in many scenarios. Single frame object detectors have recently become very powerful \cite{Redmon2017, Liu2016, Jifeng2016,Lin2017, Cao2017, xiao2018simple}. These will for each frame in a video give a list of objects in the scene and for each object its class (person, car, ...) and some representation of the object location (bounding box, keypoints, pixel mask, ...). These detections can then be connected from frame to frame into object tracks using a multi target tracking algorithm. The task of that algorithm is also to discard false detections and fill in missing detections. This paper proposes and investigates a novel framework to address this multi object tracking problem.

Network flow-based methods is a classical approach to resort to in multi target tracking \cite{Kim2015, Pirsiavash:2011:GGA:2191740.2191761, Berclaz2011, DBLP:conf/cvpr/ZhangLN08, Frossard2018}. They are computational efficient and it is often possible to guarantee a globally optimal solution. However it has been argued that  these are "very restrictive in representing motion and appearance" due to the fact that they can only contain unary and pairwise terms \cite{Kim2015}. In this paper we show how motion can be incorporated into such methods by using optical flow to observe the motion (instead of estimating it from observed positions) and introducing long range connections in the graph. The later also allows the full tracking problem to be solved with a single optimization without the need to first produce tracklets that are later combined into tracks, which is often otherwise needed \cite{DBLP:conf/eccv/RistaniSZCT16, DBLP:conf/cvpr/ZhangLN08}. Classically, network flow-based methods were derived from statistical models. However such models quickly become very complicated as they need to handle a varying number of objects which leads to the need of maximizing over likelihoods from probability-spaces of different dimensions. Such likelihoods are not directly comparable. There are solutions based on for example finite set statistics (FISST) \cite{Mahler:2007:SMI:1512927}, but these models quickly become untractable and severe approximations have to be applied to actually use them \cite{1261119}. The recent trend here is to learn such models from data instead. For example, Frossard and Urtasun \cite{Frossard2018} train a detector and a flow based tracker end-to-end. A linear program is used to maximize flow during the inference and during training one  backpropagates through this maximization.

In this paper, we present a new multi target tracking framework that learns a cost-flow weights model from data. It embeds all feasible solutions into a one dimensional feature space consisting of a score with the aim of making the score of the correct solution higher than all incorrect solutions. Then the linear program is used  during inference to efficiently search for the correct solution. We also introduce a data representation denoted generalized graph differences and show that it allows the training can be performed efficiently both in terms of training speed and data needs.

The setup proposed is similar in sprit to recent works \cite{Frossard2018, Schulter2017DeepNF}. However, they need to solve a linear program or a general convex problem respectively for each example during each step of the SGD-like optimisation, which is time consuming operations. Also, there is no need to approximate and reformulate the model as Schulter et al. \cite{Schulter2017DeepNF} does.

The small and efficient representation of generalized graph differences gives the potential for using larger graphs which is needed to fill in missing detections during, for example, occlusions by long range connections in the graphs. A key insight here is that lots of small generalized graph differences can be generated from a single annotated video sequence and be utilized as training data. This gives a good way to utilize the annotations as much as possible in order to avoid the need for extreme amounts of training data. We also show that by using average-pooling it is possible to use features for connecting detections that are derived from a varying number of feature point tracks of varying length.

The main contributions of this paper are:
\begin{itemize}
\item The ability to train cost-flow weights models without slow backproping over linear programs or general convex solvers, see Section~\ref{sec:train}.
\item A data augmentation approach to generate general graph differences for efficient training with regard to time consumption and data utilisation, see Section~\ref{sec:ggd} and Figure~\ref{fig:learning_curve}.
\item State of the art MOTA score on the hard test set of the DukeMTMCT challenge, see Section~\ref{sec:experiments} and Table~\ref{tab:dukeresult}.
\end{itemize}


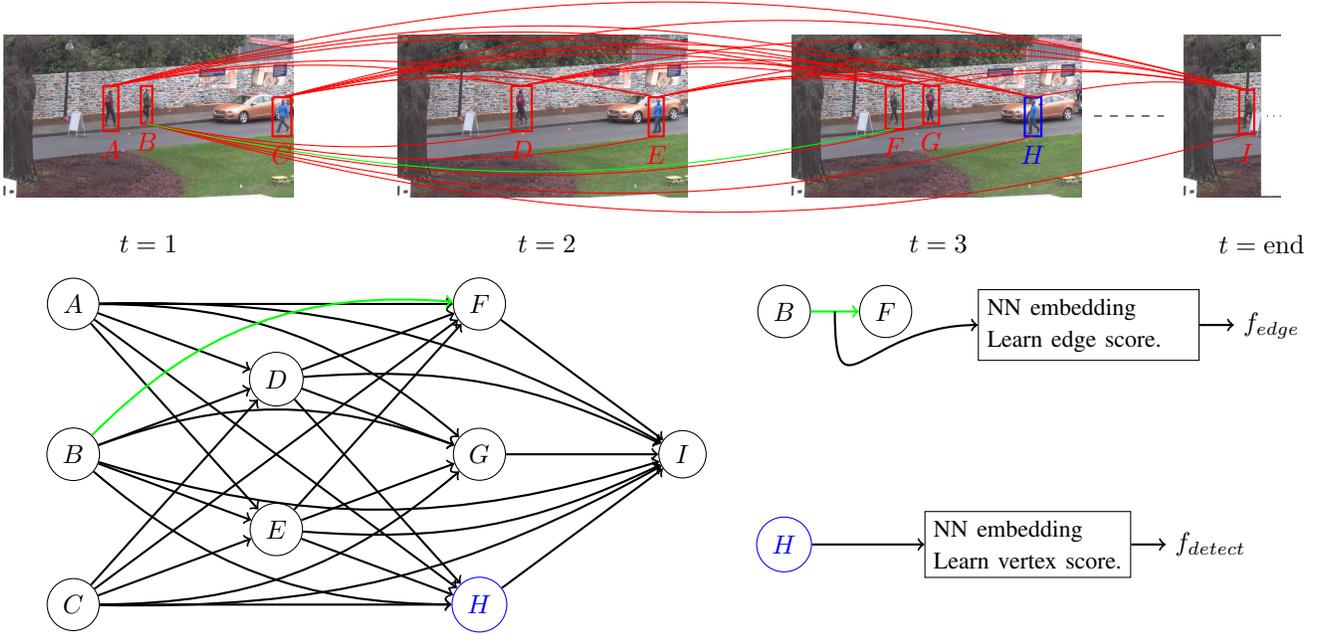
\begin{figure*}
\begin{tikzpicture}
\node[inner sep=0pt] (im1) at (0,0)
    {\includegraphics[width=.22\textwidth]{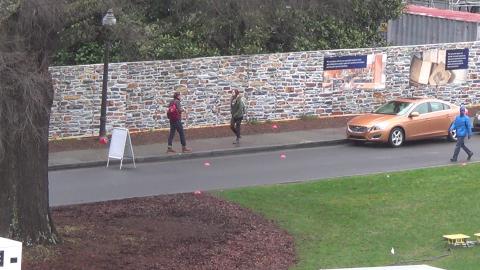}};
\node[inner sep=0pt] (im2) at (0.3\textwidth,0)
    {\includegraphics[width=.22\textwidth]{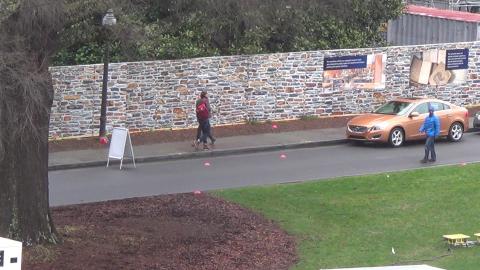}};
\node[inner sep=0pt] (im3) at (0.6\textwidth,0)
    {\includegraphics[width=.22\textwidth]{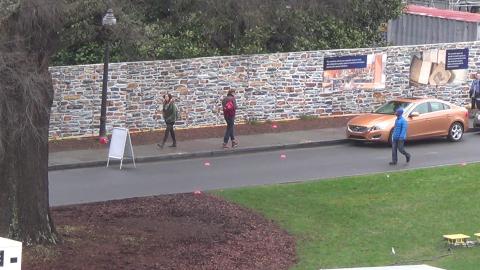}};
\node[inner sep=0pt] (im4) at (0.825\textwidth,0)
    {\includegraphics[width=.073333\textwidth]{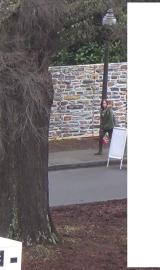}};

\pgfmathsetmacro{\xa}{-0.6}
\pgfmathsetmacro{\xb}{-0.4}
\pgfmathsetmacro{\ya}{0.4}
\pgfmathsetmacro{\yb}{-0.2}
\draw[thick,red] (\xa,\ya) -- (\xb,\ya) -- (\xb,\yb) -- (\xa,\yb) -- (\xa,\ya);
\coordinate (AU) at (-0.5,0.4);
\coordinate (AD) at (-0.5,-0.2);

\pgfmathsetmacro{\xa}{-0.1}
\pgfmathsetmacro{\xb}{0.05}
\pgfmathsetmacro{\ya}{0.4}
\pgfmathsetmacro{\yb}{-0.1}
\draw[thick,red] (\xa,\ya) -- (\xb,\ya) -- (\xb,\yb) -- (\xa,\yb) -- (\xa,\ya);
\coordinate (BU) at (-0.0250,-0.2);
\coordinate (BD) at (-0.0250,-0.1);

\pgfmathsetmacro{\xa}{1.65}
\pgfmathsetmacro{\xb}{1.9}
\pgfmathsetmacro{\ya}{0.25}
\pgfmathsetmacro{\yb}{-0.27}
\draw[thick,red] (\xa,\ya) -- (\xb,\ya) -- (\xb,\yb) -- (\xa,\yb) -- (\xa,\ya);
\coordinate (CU) at (1.7750,0.25);
\coordinate (CD) at (1.7750,-0.27);

\pgfmathsetmacro{\xa}{-0.65+5.475}
\pgfmathsetmacro{\xb}{-0.38+5.475}
\pgfmathsetmacro{\ya}{0.4}
\pgfmathsetmacro{\yb}{-0.2}
\draw[thick,red] (\xa,\ya) -- (\xb,\ya) -- (\xb,\yb) -- (\xa,\yb) -- (\xa,\ya);
\coordinate (DU) at (4.96,0.4);
\coordinate (DD) at (4.96,-0.2);

\pgfmathsetmacro{\xa}{1.65+5}
\pgfmathsetmacro{\xb}{1.85+5}
\pgfmathsetmacro{\ya}{0.25}
\pgfmathsetmacro{\yb}{-0.27}
\draw[thick,red] (\xa,\ya) -- (\xb,\ya) -- (\xb,\yb) -- (\xa,\yb) -- (\xa,\ya);
\coordinate (EU) at (1.75+5,0.25);
\coordinate (ED) at (1.75+5,-0.27);

\pgfmathsetmacro{\xa}{-0.6+10.4}
\pgfmathsetmacro{\xb}{-0.4+10.43}
\pgfmathsetmacro{\ya}{0.4}
\pgfmathsetmacro{\yb}{-0.17}
\draw[thick,red] (\xa,\ya) -- (\xb,\ya) -- (\xb,\yb) -- (\xa,\yb) -- (\xa,\ya);
\coordinate (FU) at (9.9150,0.4);
\coordinate (FD) at (9.9150,-0.17);

\pgfmathsetmacro{\xa}{-0.1+10.4}
\pgfmathsetmacro{\xb}{0.05+10.46}
\pgfmathsetmacro{\ya}{0.4}
\pgfmathsetmacro{\yb}{-0.1}
\draw[thick,red] (\xa,\ya) -- (\xb,\ya) -- (\xb,\yb) -- (\xa,\yb) -- (\xa,\ya);
\coordinate (GU) at (10.4050,0.4);
\coordinate (GD) at (10.4050,-0.1);

\pgfmathsetmacro{\xa}{1.65+10}
\pgfmathsetmacro{\xb}{1.87+10}
\pgfmathsetmacro{\ya}{0.25}
\pgfmathsetmacro{\yb}{-0.27}
\draw[thick,blue] (\xa,\ya) -- (\xb,\ya) -- (\xb,\yb) -- (\xa,\yb) -- (\xa,\ya);
\coordinate (HU) at (11.7600,0.25);
\coordinate (HD) at (11.7600,-0.27);

\pgfmathsetmacro{\xa}{-0.6+15.1}
\pgfmathsetmacro{\xb}{-0.4+15.1}
\pgfmathsetmacro{\ya}{0.35}
\pgfmathsetmacro{\yb}{-0.24}
\draw[thick,red] (\xa,\ya) -- (\xb,\ya) -- (\xb,\yb) -- (\xa,\yb) -- (\xa,\ya);
\coordinate (IU) at (14.6,0.35);
\coordinate (ID) at (14.6,-0.24);

\node at (0,-1.7) {$t=1$};
\node at (5.3,-1.7) {$t=2$};
\node at (10.5,-1.7) {$t=3$};
\node at (14.8,-1.7) {$t=\mathrm{end}$};

\draw [red] (AU) to[out=10,in=170] (DU);
\draw [red] (AU) to[out=11,in=169] (EU);
\draw [red] (AU) to[out=12,in=168] (FU);
\draw [red] (AU) to[out=13,in=167] (GU);
\draw [red] (AU) to[out=14,in=166] (HU);
\draw [red] (AU) to[out=15,in=165] (IU);

\draw [red] (BD) to[out=-10,in=190] (DD);
\draw [red] (BD) to[out=-11,in=191] (ED);
\draw [red] (BD) to[out=-13,in=193] (GD);
\draw [red] (BD) to[out=-14,in=194] (HD);
\draw [red] (BD) to[out=-15,in=195] (ID);

\draw [red] (CU) to[out=10,in=170] (DU);
\draw [red] (CU) to[out=11,in=169] (EU);
\draw [red] (CU) to[out=12,in=168] (FU);
\draw [red] (CU) to[out=13,in=167] (GU);
\draw [red] (CU) to[out=14,in=166] (HU);
\draw [red] (CU) to[out=15,in=165] (IU);

\draw [red] (DU) to[out=10,in=170] (FU);
\draw [red] (DU) to[out=11,in=169] (GU);
\draw [red] (DU) to[out=12,in=168] (HU);
\draw [red] (DU) to[out=13,in=167] (IU);

\draw [red] (EU) to[out=10,in=170] (FU);
\draw [red] (EU) to[out=11,in=169] (GU);
\draw [red] (EU) to[out=12,in=168] (HU);
\draw [red] (EU) to[out=13,in=167] (IU);

\draw [red] (FU) to[out=10,in=170] (IU);
\draw [red] (GU) to[out=10,in=170] (IU);
\draw [red] (HU) to[out=10,in=170] (IU);

\draw [green] (BD) to[out=-12,in=192] (FD);

\coordinate (a) at (1.87+10.7,0);
\coordinate (b) at (-0.6+14.1,0);
\draw[dashed] (a) -- (b);

\coordinate (a) at (1.87+13,0);
\coordinate (b) at (1.87+13+0.25,0);
\draw[dotted] (a) -- (b);

\node[below,red] at (AD) {$A$};
\node[below,red] at (BD) {$B$};
\node[below,red] at (CD) {$C$};
\node[below,red] at (DD) {$D$};
\node[below,red] at (ED) {$E$};
\node[below,red] at (FD) {$F$};
\node[below,red] at (GD) {$G$};
\node[below,red] at (ID) {$I$};
\node[below,blue] at (HD) {$H$};

\def \offx {-1}
\def \offy {-2.5}
\def \dy {-2}
\def \dx {2.7}

\node[draw, circle] (A) at (\offx,\offy) {$A$};
\node[draw, circle] (B) at (\offx,\offy+\dy) {$B$};
\node[draw, circle] (C) at (\offx,\offy+2*\dy) {$C$};

\node[draw, circle] (D) at (\offx+\dx,\offy+\dy/2) {$D$};
\node[draw, circle] (E) at (\offx+\dx,\offy+\dy*3/2) {$E$};

\node[draw, circle] (F) at (\offx+2*\dx,\offy) {$F$};
\node[draw, circle] (G) at (\offx+2*\dx,\offy+\dy) {$G$};
\node[draw, circle, blue] (H) at (\offx+2*\dx,\offy+2*\dy) {$H$};

\node[draw, circle] (I) at (\offx+3*\dx,\offy+\dy) {$I$};

\path[->,draw,thick]
    (A) edge node {$$} (D)  ;

\path[->,draw,thick]
    (A) edge node {$$} (E)  ;

\path[->,draw,thick]
    (A) edge node {$$} (F)  ;

\path[->,draw,thick]
    (A) edge [bend left=20]  node {$$} (G)  ;

\path[->,draw,thick]
    (A) edge node {$$} (H)  ;

\path[->,draw,thick]
    (A) edge [bend left=15]  node {$$} (I)  ;

\path[->,draw,thick]
    (B) edge node {$$} (D)  ;

\path[->,draw,thick]
    (B) edge node {$$} (E)  ;

\path[->,draw,thick]
    (B) edge [bend left=20]  node {$$} (G)  ;

\path[->,draw,thick]
    (B) edge  [bend right=20]  node {$$} (H)  ;

\path[->,draw,thick]
    (B) edge [bend right=17]  node {$$} (I)  ;

\path[->,draw,thick]
    (C) edge node {$$} (D)  ;

\path[->,draw,thick]
    (C) edge node {$$} (E)  ;

\path[->,draw,thick]
    (C) edge node {$$} (F)  ;

\path[->,draw,thick]
    (C) edge [bend right=20]  node {$$} (G)  ;

\path[->,draw,thick]
    (C) edge node {$$} (H)  ;

\path[->,draw,thick]
    (C) edge [bend right=15]  node {$$} (I)  ;

\path[->,draw,thick]
    (D) edge node {$$} (F)  ;

\path[->,draw,thick]
    (D) edge  node {$$} (G)  ;

\path[->,draw,thick]
    (D) edge node {$$} (H)  ;

\path[->,draw,thick]
    (D) edge [bend left=15]  node {$$} (I)  ;

\path[->,draw,thick]
    (E) edge node {$$} (F)  ;

\path[->,draw,thick]
    (E) edge  node {$$} (G)  ;

\path[->,draw,thick]
    (E) edge node {$$} (H)  ;

\path[->,draw,thick]
    (E) edge [bend right=15]  node {$$} (I)  ;

\path[->,draw,thick]
    (F) edge node {$$} (I)  ;

\path[->,draw,thick]
    (G) edge node {$$} (I)  ;

\path[->,draw,thick]
    (H) edge node {$$} (I)  ;

\path[->,draw,thick,green]
    (B) edge [bend left=25]  node {$$} (F)  ;

\node[draw, circle] (BB) at (\offx+3.5*\dx,\offy-0.1) {$B$};
\node[draw, circle] (FF) at (\offx+4*\dx,\offy-0.1) {$F$};
\path[->,draw,thick,green]
    (BB) edge node {$$} (FF)  ;


\node[draw, circle,blue] (HH) at (\offx+3.5*\dx,\offy+\dy*3.2/2) {$H$};

\node[draw, rectangle, text width=2.7cm,align=left] (NNE) at (\offx+5*\dx,\offy+0.14*\dy) {\small NN embedding\\Learn edge score.};
\node(NNES) at (\offx+5.9*\dx,\offy+0.14*\dy) {$f_{edge}$};
\path[->,draw,thick]
    (NNE) edge  node {$$} (NNES)  ;

\node[draw, rectangle, text width=2.5cm,align=left] (NNV) at (\offx+4.7*\dx,\offy+1.6*\dy) {\small NN embedding\\Learn vertex score.};
\node(NNVS) at (\offx+5.6*\dx,\offy+1.6*\dy) {$f_{detect}$};
\path[->,draw,thick]
    (NNV) edge  node {$$} (NNVS)  ;


\path[->,draw,thick] (\offx+3.75*\dx,\offy-0.1) to [out=-90,in=180,looseness=2] (NNE.west)  ;
\path[->,draw,thick] (HH) edge  node {$$} (NNV)  ;

\end{tikzpicture}
\caption{Concept of proposed method to address tracking with a graph and learning mapping for edges and vertices.}
\label{fig:overview}
\end{figure*}

\section{Proposed Tracking Algorithm}

Here the proposed algorithm will be formulated as a constrained mathematical optimisation problem. It can then be solved using either network flow algorithms or more general linear programming. To keep the formulation simple, only a single object class is considered. But generalizing to multiple classes is straight forward.

\subsection{Basic graph formulation}

The basic idea behind the algorithm is to build a graph with object detections as vertices and use sparse optical flow feature point tracks, KLT-tracks \cite{Bouguet00pyramidalimplementation}, to connect these vertices with edges. Then a flow capacity of one is assigned to each edge and a network flow problem is solved. The solution will have a positive flow of one between detections that belong to the same object, see Figure~\ref{fig:overview}.

The input to the algorithm is a set of detections,
\begin{equation}
	V = \left\{ v_0, v_1, \cdots v_{\left|V\right|}\right\} ,
\end{equation}
produced by an object detector. Each detection, $v_k=\left(t_k, L_k, c_k\right)$ consists of a frame number, $t_k$, a location, $L_k$ and a confidence $c_k$. The location represents which pixels in the image the object covers. It can be defined as a bounding-box or as a maximum distance to some keypoints or as a pixel level segmentation. The only assumption is that there exists an indicator function which can tell if a pixel, $\left(x, y\right)$ is located on the object (or at least close by in case of less precise representations). That indicator function will be denoted $\left(x, y\right) \in L_k$.

In addition to the detections there is also KLT-tracks consisting of a set of point tracks
\begin{equation}
	P = \left\{ P_0, P_1, \cdots, P_{\left|P\right|}\right\} ,
\end{equation}
where $P_i = \left(p_{i,0}, p_{i,1}, \cdots p_{i,\left|P_i\right|}\right)$ and $p_{i,j} = \left(t_{i,j}, x_{i,j}, y_{i,j}, c_{i,j}\right)$. Here $t_{i,j}$ is the global frame number and $\left(x_{i,j}, y_{i,j}\right)$ is the pixel location of the KLT-track in that frame and $c_{i,j}$ is a confidence. The confidence used here is the negated L1 distance between a small patch centered around the point in frame $t_{i,j}$ and $t_{i,j}$-1.

Each KLT-track will connect the detections it intersects into a sequences of detections. Each such sequence form one object track hypothesis. All of those hypothesis will be combined into edges in a graph representing different possible object tracks.

To formalize, a set $A_i$ is introduced, that contains all detections intersecting the feature point track $P_i$,
\begin{equation}
	A_i =
	\left\{v_k \left| t_{i,j}=t_k, \left(x_{i,j}, y_{i,j}\right) \in L_k \text{\quad for some $j$}\right.\right\} .
\end{equation}
Then a graph is formed where the detections, $v_k$, are vertices and edges between the vertices are produced from neighbouring detections within each of the $A_i$ tracks. Note that the distance between neighbouring detections in $A_i$ might be several frames as the feature points can be tracked even if there are no detections. A neighbouring radius of $r_\text{neighbours}$ is used. That is, a detection is considered to be neighbour with the $r_\text{neighbours}$ preceding and $r_\text{neighbours}$ following detections. In order to avoid connecting  distant detection a threshold, $t_\text{max}$, is introduced to discard such edges. That is, two detections, $v_{k_1}$ and $v_{k_2}$ are not considered neighbours if $\left|t_{k_2} - t_{k_1}\right| > t_\text{max}$. Also note that in the case of overlapping detections, $A_i$ might contain two (or more) detections for the same frame. These detections are not considered neighbours to each other. Instead their neighbouring detections will have multiple incoming or outgoing edges.
Formally, let $\left(v_{k_1}, v_{k_2}\right) \in N_i$ denote that $v_{k_1}$ and $v_{k_2}$ are neighbours in $A_i$ according to the neighbouring structure described above. Then
there is a set of directed edges,

\begin{equation}
	E = \left\{
	\left(v_{k_1}, v_{k_2}\right)
	\left| t_{k_2} > t_{k_1}, \left(v_{k_1}, v_{k_2}\right) \in N_i
	\text{\quad for some $i$}
	\right.\right\} .
\end{equation}

Each edge is weighted with a weight function $f_\text{edge}$
that depend on all the KLT-tracks between the detections $v_{k_1}$ and $v_{k_2}$, $P_{k_1,k_2} = \left\{P_i \left| \right. p_{i,j_1} \in L_{k_1}, p_{i,j_2} \in L_{k_2} \text{\quad for some $j_1$, $j_2$}\right\}$. The vertexes are also weighted with a weight function $f_\text{detect}$ that depend on the detection $v_k$. These are learned from training data, see Section~\ref{sec:train}.


\subsection{Long range connections}
\label{sec:long}
To allow objects to occlude each other, long range connections can be added to the graph. The problem is that during an occlusion a lot of feature point tracks will jump from one object to the other, which means that the feature point tracks are not reliably in such situations. In order to address this issue, the common used linear motion model is utilized in this setup \cite{Luo14}. Other motion models could be considered, and would fit into the framework with minimal modifications, but that pursuit is out of the scope in this paper. Therefore, a velocity, $w_{i,k}^\text{pre}$ is also estimated for each KLT-track, $P_i$, intersecting the detection $v_k$. It is produced
by fitting a line to the $n_\text{velest}$ most recent positions preceding $t_k$ of that KLT-track.
Using this velocity the location can be projected into the $n_\text{project}$ closest future frames, and connections made to detections there. The weights of such connections will depend both on how well the future detection matches the predicted location and on how well the estimated velocities match. This kind of edges can skip over problematic situations entirely and instead match velocity and position of incoming and outgoing tracks. The velocity of the outgoing detections, $w_{i,k}^\text{post}$, is calculated from the $n_\text{velest}$ KLT-track positions directly following the detection time $t_k$. This way the incoming velocity is estimated prior to the occlusion and the outgoing velocity is estimated after the occlusion. That means that neither of them should be affect too much by confusing KLT-tracks jumping target  during the occlusion.

A set of long connections, $C_{k_1,k_2}$, connecting $v_{k_1}$ and $v_{k_2}$ will be formed, with one connection for each KLT-track intersecting $v_{k_1}$ that started more than $n_\text{velest}$ before $t_{k_1}$. Each of these connection will be based on
different velocity estimates, $w_{i,k_1}^\text{pre}$. That is

\begin{equation}
C_{k_1,k_2} = \left\{ w_{i,k_1}^\text{pre} \left|
p_{i,j} \in L_{k_1}, \text{\quad for some $j \geq n_\text{velest}$}
\right.\right\} .
\end{equation}

Now the edge weight function, $f_\text{edge}$ depend on both the KLT-tracks and the long connections,
$f_\text{edge} \left(P_{k_1,k_2}, C_{k_1,k_2}, v_{k_1}, v_{k_2}\right)$.

\subsection{Network flow}

Using the constructed graph, the multi target tracking problem can be formulated as a network flow problem. Indicator variables, $\hat v_k \in \left\{0, 1\right\}$, are introduced that indicates whether each detection is a true positive or a false positive. Also, indicator variables, $\hat e_{k_1,k_2} \in \left\{0, 1\right\}$, for the edges are introduced. The edges indicate that the two detections they connect are adjacent connections of the same track. One of the features used to form the edge weights will be the temporal difference of the detections, which allows for a penalty for missing detections to be learnt. Finally, $\hat f_k, \hat l_k \in \left\{0, 1\right\}$ are introduced to indicate that $v_k$ is the first, $\hat f_k$, and/or the last, $\hat l_k$, detection of a track. By denoting the combination of these indicators $x=\left(v_1, f_1, l_1, e_{1,2}, v_2, \cdots \right)$, the score, $f_\text{score}\left(x\right)=$
\begin{multline}
	\sum_k \hat f_k s_\text{entry} +
	\sum_k \hat v_k f_\text{detect}\left(v_k\right) + \\
	+ \sum_{k_1,k_2} \hat e_{k_1,k_2}
	f_\text{edge} \left(P_{k_1,k_2}, C_{k_1,k_2}, v_{k_1}, v_{k_2}\right)
	\label{eq:score}
\end{multline}
can be optimized to find the best solution to the tracking problem.
Here, $s_\text{entry}$ is a negative number efficiently becoming a threshold on the total track score for a track not to be considered noise.

Constraints have to be introduced to ensure that it is a proper solution in the sense that each detection only belongs to a single track and that unconnected detections are considered false positives. These are the flow constraints with flow variables both on edges and on vertices \cite{Frossard2018}. It ensures that the  outgoing flow of each vertex is the same as the incoming flow and equal to the flow variable of the vertex. The constraints are
\begin{equation}
	\hat v_k = \hat f_k + \sum_{k_1} \hat e_{k_1,k} = \hat l_k + \sum_{k_2} \hat e_{k,k_2} .
	\label{eq:constraints}
\end{equation}
The solutions, $x$, that fulfills this equation are considered feasible solutions and the set of them is denoted $S$, which allows the tracking problem to be expressed as
\begin{equation}
	\argmax_{x \in S} f_\text{score}\left(x\right) .
	\label{eq:motaopt}
\end{equation}

\subsection{Optimization}

The multi target tracking problem can be formulated as the maximisation in Equation~\ref{eq:motaopt}. It can be solved using a linear program. This is guaranteed to result in a integer solution as it exhibits the total unimodularity property \cite{Berclaz2011}. A more efficient way is to convert the linear program into a classical network cost flow problem \cite{DBLP:conf/cvpr/ZhangLN08} by replacing each vertex with two vertexes connected with a single edge with the original vertex weight as the edge weight and placing all incoming edges on one of these vertexes and all outgoing edges on the other. This network flow problem can then be solved using Bellmann-Ford \cite{Bertsekas:1992:DN:121104} or Successive Shortest Paths \cite{Ahuja:1993:NFT:137406}. Yet another alternative is to use K-shortest paths \cite{Berclaz2011}.

\section{Parameter learning}
\label{sec:train}

\begin{figure}
\resizebox{\columnwidth}{!}{%
\begin{tikzpicture}
\tikzset{
  pics/nn/.style n args={4}{
    code = { %
    \small
          \node[draw,fill=green!50,thick,circle,label={$x_1$}] (A) at (#1,#2) {$ $};
          \node[draw,fill=blue!50,thick,circle,label={$h_{1,1}$}] (B) at (#1+#3,#2) {$ $};
          \node[draw,fill=blue!50,thick,circle,label={$h_{1,2}$}] (C) at (#1+2*#3,#2) {$ $};
          \node (D) at (#1+2.7*#3,#2) {$\ldots$};
          \node[draw,fill=blue!50,thick,circle, label={$h_{1,n_{layers}}$}] (E) at (#1+3.4*#3,#2) {$ $};
          \node[draw,fill=red!50,thick,circle,label={$y_1$}] (F) at (#1+4.4*#3,#2) {$ $};

          \node[draw,fill=green!50,thick,circle,label={$x_2$}] (AA) at (#1,#2-#4) {$ $};
          \node[draw,fill=blue!50,thick,circle,label={$h_{2,1}$}] (BB) at (#1+#3,#2-#4) {$ $};
          \node[draw,fill=blue!50,thick,circle,label={$h_{2,2}$}] (CC) at (#1+2*#3,#2-#4) {$ $};
          \node (DD) at (#1+2.7*#3,#2-#4) {$\ldots$};
          \node[draw,fill=blue!50,thick,circle, label={$h_{2,n_{layers}}$}] (EE) at (#1+3.4*#3,#2-#4) {$ $};
          \node[draw,fill=red!50,thick,circle,label={$y_2$}] (FF) at (#1+4.4*#3,#2-#4) {$ $};

          \node (AAA) at (#1,#2-2*#4) {$\vdots $};
          \node (BBB) at (#1+#3,#2-2*#4) {$\vdots $};
          \node (CCC) at (#1+2*#3,#2-2*#4) {$\vdots $};
          \node (DDD) at (#1+2.7*#3,#2-2*#4) {$\ldots$};
          \node (EEE) at (#1+3.4*#3,#2-2*#4) {$\vdots $};
          \node (FFF) at (#1+4.4*#3,#2-2*#4) {$\vdots $};

          \node[draw,fill=green!50,thick,circle,label={$x_{n_{input}}$}] (AAAA) at (#1,#2-3*#4) {$ $};
          \node[draw,fill=blue!50,thick,circle,label={$h_{n_{hidden},1}$}] (BBBB) at (#1+#3,#2-3*#4) {$ $};
          \node[draw,fill=blue!50,thick,circle,label={$h_{n_{hidden},2}$}] (CCCC) at (#1+2*#3,#2-3*#4) {$ $};
          \node (DDDD) at (#1+2.7*#3,#2-3*#4) {$\ldots$};
          \node[draw,fill=blue!50,thick,circle, label={$h_{n_{hidden},n_{layers}}$}] (EEEE) at (#1+3.4*#3,#2-3*#4) {$ $};
          \node[draw,fill=red!50,thick,circle,label={$y_{n_{output}}$}] (FFFF) at (#1+4.4*#3,#2-3*#4) {$ $};

          \node (O) at (#1+5*#3,0) {$\, $};

          \draw[->,gray] (A) -- (B);
          \draw[->,gray] (A) -- (BB);
          \draw[->,gray] (A) -- (BBBB);
          \draw[->,gray] (AA) -- (B);
          \draw[->,gray] (AA) -- (BB);
          \draw[->,gray] (AA) -- (BBBB);
          \draw[->,gray] (AAAA) -- (B);
          \draw[->,gray] (AAAA) -- (BB);
          \draw[->,gray] (AAAA) -- (BBBB);

          \draw[->,gray] (B) -- (C);
          \draw[->,gray] (B) -- (CC);
          \draw[->,gray] (B) -- (CCCC);
          \draw[->,gray] (BB) -- (C);
          \draw[->,gray] (BB) -- (CC);
          \draw[->,gray] (BB) -- (CCCC);
          \draw[->,gray] (BBBB) -- (C);
          \draw[->,gray] (BBBB) -- (CC);
          \draw[->,gray] (BBBB) -- (CCCC);

          \draw[->,gray] (E) -- (F);
          \draw[->,gray] (E) -- (FF);
          \draw[->,gray] (E) -- (FFFF);
          \draw[->,gray] (EE) -- (F);
          \draw[->,gray] (EE) -- (FF);
          \draw[->,gray] (EE) -- (FFFF);
          \draw[->,gray] (EEEE) -- (F);
          \draw[->,gray] (EEEE) -- (FF);
          \draw[->,gray] (EEEE) -- (FFFF);

          \normalsize

    }
  }
}

\pic{nn={0}{0}{1.75}{1}};

\end{tikzpicture}
}
\caption{Generic fully connect neural network used in framework.}
\label{fig:nnblock}
\end{figure}
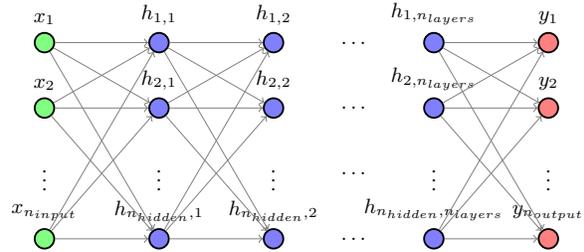

The tracking model in the previous section contains some functions that needs to be learned from annotated training sequences. These sequences are training examples consisting of short videos fully annotated with multi object tracking ground truth. Fully connected neural networks will be used as basic blocks to construct these functions. These blocks are parameterised with two parameters only, the number of layers and the number of features. All layers have the same number of features, see Figure~\ref{fig:nnblock}.

\subsection{Model architecture}
The parameters that needs to be learned are the scalar $s_\text{entry}$ and the embedded parameters in the functions
$f_\text{detect}\left(v_k\right)$ and
$f_\text{edge}\left(P_{k_1,k_2}, C_{k_1,k_2}, v_{k_1}, v_{k_2}\right)$. These functions will be implemented as neural networks and it is the parameters of those networks that needs to learned together with $s_\text{entry}$.

The detection score, $f_\text{detect}\left(v_k\right)$, is a scalar valued function that depend on features extracted from the detection, $v_k$. The features used are
\begin{itemize}
\item The detection confidence, $c_k$.
\item The maximum IoU between the detection $v_k$ and any other detection in the same frame.
\item The maximum IoA (intersection over area of $v_k$) between the detection $v_k$ and any other detection in the same frame.
\end{itemize}
The detection score function, $f_\text{detect}$, will be realized as a small neural network with three inputs and one output, the detection score. The network has $n_\text{detlayers}$ fully connected hidden layers with $n_\text{detfeat}$ features each.

\begin{figure*}
\centering
\includegraphics[width=0.8\textwidth]{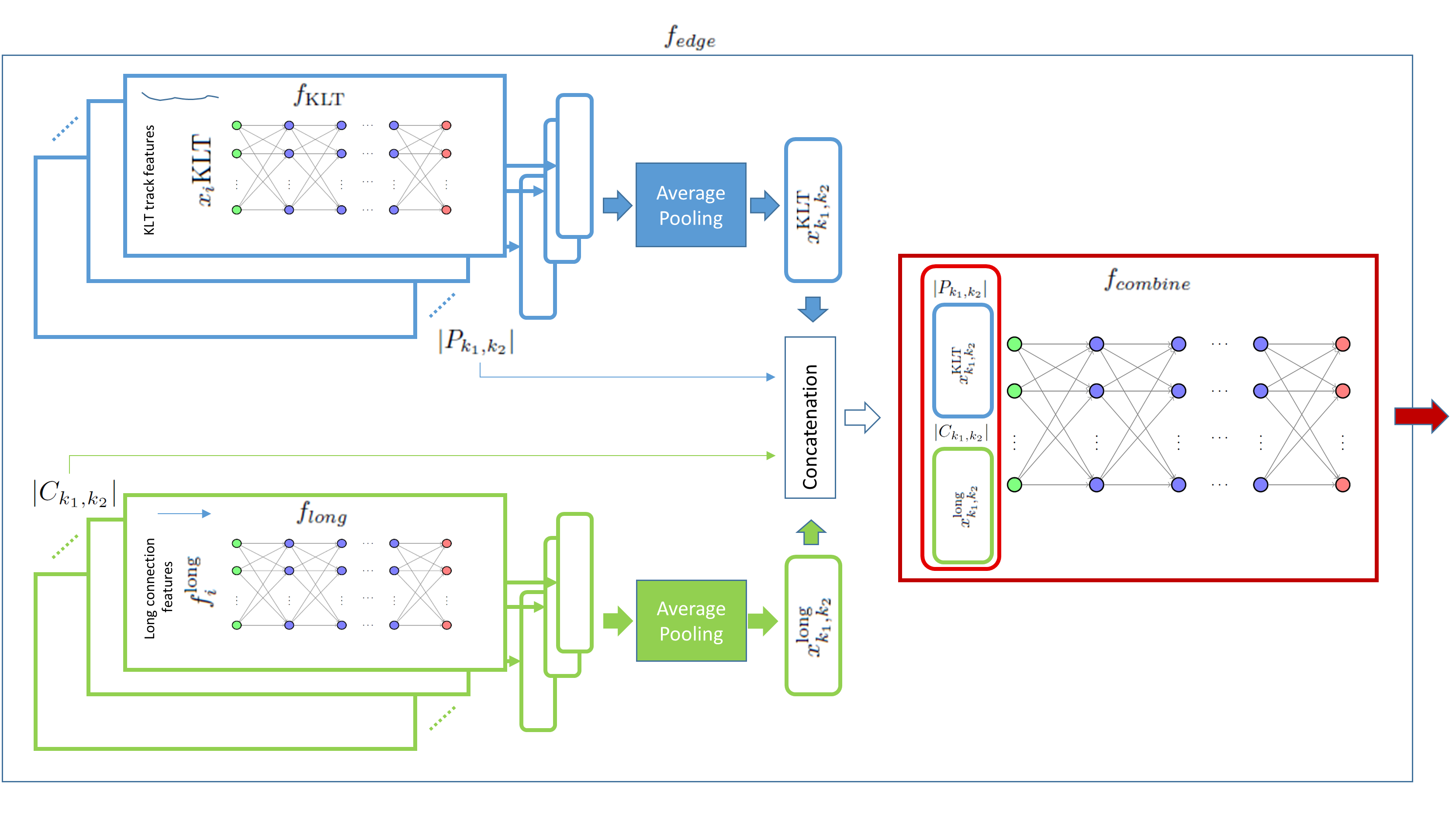}
\caption{The architecture of $f_\text{edge}\left(P_{k_1, k_2}, C_{k_1,k_2}, v_{k_1}, v_{k_2}\right)$.}
\label{fig:fedge}
\end{figure*}

\begin{figure}
\centering
\includegraphics[width=0.8\columnwidth]{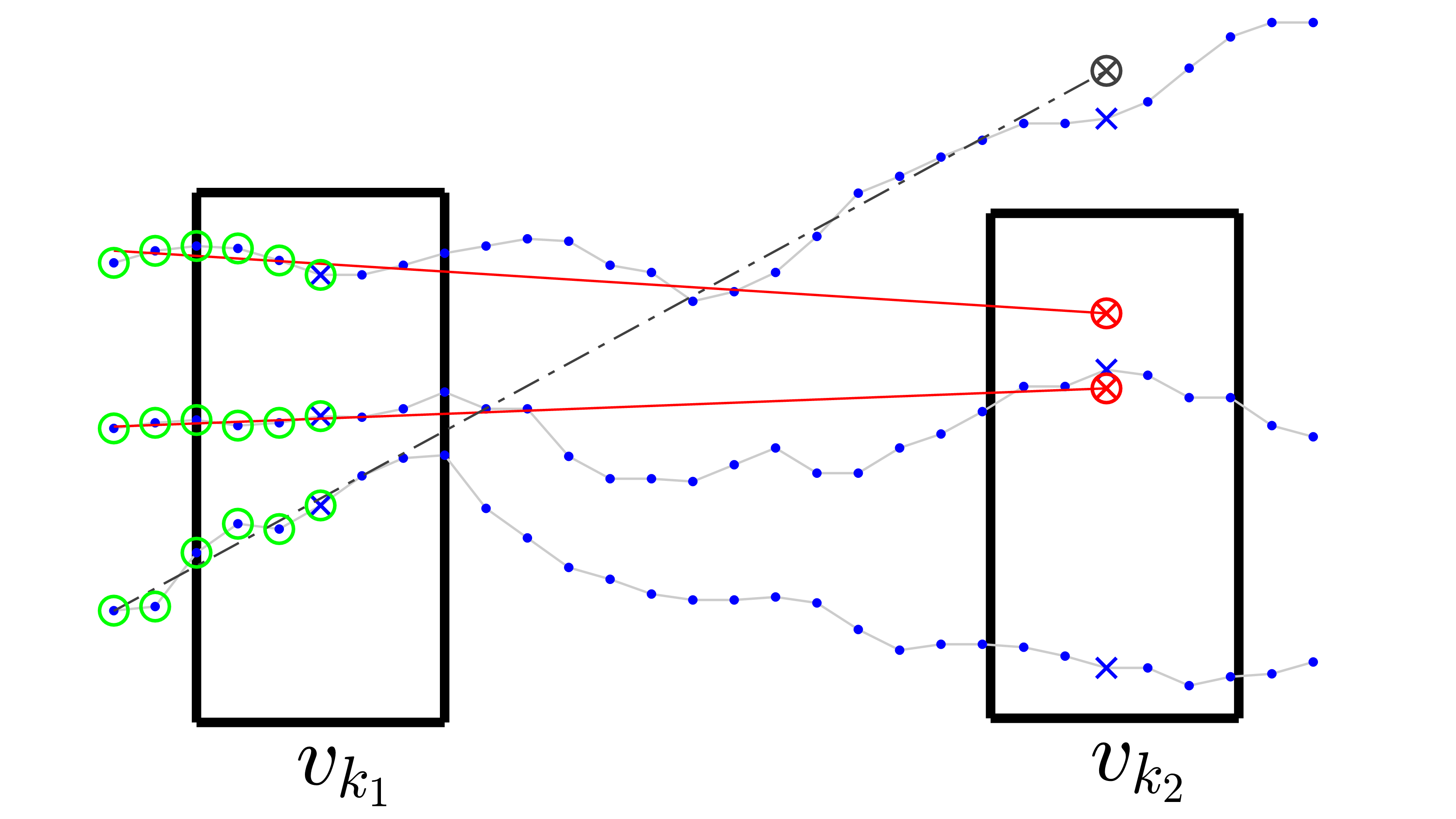}
\caption{A pair of detections, $\left(v_{k_1}, v_{k_2}\right)$ (black boxes) connected with
$\left|P_{v_{k_1}, v_{k_2}}\right|=2$ KLT-tracks (blue dots) and $\left|C_{v_{k_1}, v_{k_2}}\right|=2$ long connections (red lines) estimated from $n_\text{velest}=6$ positions (green circles).
}
\label{fig:tracksv1v2}
\end{figure}

The edge score, $f_\text{edge}\left(P_{k_1, k_2}, C_{k_1,k_2}, v_{k_1}, v_{k_2}\right)$ is more complicated and an overview of it is shown in Figure~\ref{fig:fedge}. It depend on
all KLT-track connections, $P_{k_1, k_2}$, and all long connections, $C_{k_1,k_2}$, connecting the detections $v_{k_1}$ and  $v_{k_2}$, see example in Figure~\ref{fig:tracksv1v2}.
The number of such connections will vary from vertex to vertex, as will the number of positions in the KLT-tracks. To handle that each KLT-track, $P_i \in P_{k_1, k_2}$ is converted into a fixed length feature vector, $x_i^\text{KLT}$, consisting of the features
\begin{itemize}
\item Temporal distance, $t_{k_2} - t_{k_1}$.
\item Minimum confidence, $\min_j c_{i,j}$.
\item The intersection over union between $v_{k_2}$ and $v_{k_1}$ translated according to the motion of the KLT-track $P_i$.
\item A normalized trajectory shape consisting of $P_i$ translated to place $p_{i,j_1}$ (for $t_{i,j_1}=t_{k_1}$) at origo and then linearly interpolated into $n_\text{linpkt}$ points placed uniformly spaced between $t_{i,j_1}$ and $t_{i, j_2}$.
\end{itemize}

These feature vectors are processed, one by one, by a neural network, $f_\text{KLT}\left(x_i^\text{KLT}\right)$, with $n_\text{kltlayers}$ fully connected layers with $n_\text{kltfeat}$ features each. That produces one feature vector for each KLT-track. They are then combined using average-pooling to form a single fixed length feature vector,
\begin{equation}
	x^\text{KLT}_{k_1, k_2} = \frac{1}{\left|P_{k_1, k_2}\right|}
	\sum_{i\left|P_i \in P_{k_1, k_2}\right.}
	f_\text{KLT}\left(x_i^\text{KLT}\right).
\end{equation}
This allows the varying number of KLT-tracks to be processes by a network construction with a fixed number of parameters and produce a feature vector of fixed length. Training this construction is similar to training a normal neural network while varying the batch size.

In a similar fashion, the long connections, $w_{i,k_1}^\text{pre} \in C_{k_1,k_2}$ are converted to fixed length feature vectors, $x_i^\text{long}$,  with the features
\begin{itemize}
\item Temporal distance, $t_{k_2} - t_{k_1}$.
\item The intersection over union between $v_{k_2}$ and $v_{k_1}$ translated according to the predicted velocity, $w_{i,k}^\text{pre}$.
\item The predicted velocity, $w_{i,k_1}^\text{pre}$.
\item The median post velocity of $v_{k_2}$, $\median_i \left(w_{i,k_2}^\text{post}\right)$.
\end{itemize}
These feature vectors are processed, one by one, by a neural network, $f_\text{long}\left(x_i^\text{long}\right)$, with $n_\text{longlayers}$ fully connected layers with $n_\text{longfeat}$ features each, and averaged
\begin{equation}
	x^\text{long}_{k_1, k_2} = \frac{1}{\left|C_{k_1, k_2}\right|}
	\sum_{i\left|w_{i,k_1}^\text{pre} \in C_{k_1, k_2}\right.}
	f_\text{long}\left(x_i^\text{long}\right) .
\end{equation}

The feature vectors, $x^\text{klt}_{k_1, k_2}$ and $x^\text{long}_{k_1, k_2}$ are then concatenated and extended with the number of KLT-tracks, $\left|P_{k_1, k_2}\right|$ and the number of long connections, $\left|C_{k_1, k_2}\right|$ and passed to a final neural network, $f_\text{combine}\left(x^\text{klt}_{k_1, k_2}, x^\text{long}_{k_1, k_2}, \left|P_{k_1, k_2}\right|, \left|C_{k_1, k_2}\right|\right)$. This network has $n_\text{combinelayers}$ fully connected hidden layers with $n_\text{combinefeat}$ features each, and a single output, the edge feature $f_\text{edge}\left(P_{k_1,k_2}, C_{k_1,k_2}, v_{k_1}, v_{k_2}\right)$.

\subsection{Loss function}
The multi target tracking optimization problem, Equation~\ref{eq:motaopt}, can, in theory, be solved by enumerating all feasible solutions, $x\in S$ (Equation~\ref{eq:constraints}), and picking the one that maximizes $f_\text{score}\left(x\right)$ (Equation~\ref{eq:score}).
The score function, $f_\text{score}\left(x\right)$, is a linear combination of the outputs of several invocations of the neural networks defined above. That means the entire function, $f_\text{score}\left(x\right)$, is differentiable and can be learned using for example SGD.


The function can be seen as an embedding that embeds feasible solutions, $x$, into a one-dimensional feature space with the one property that better solution should have higher score.

The embedding can be learnt from training data consisting of ordered pairs of feasible solutions, $\left(x^*, x\right)$ where $x^*$ is the correct globally optimal solution and $x$ is any other feasible solution. Details of how $x$ is created in practice will be discussed later, but in general consider it to be a modification of $x^*$. The learning is achieved using the ranking loss
\begin{equation}
	-\log \sigma \left(
	    f_\text{score}\left(x^*\right) - f_\text{score}\left(x\right)
	\right) ,
	\label{eq:loss}
\end{equation}
where $ \sigma $ is the sigmoid function. It is used here to suppress large differences as the only property from the pair of interest is that
$f_\text{score}\left(x^*\right)$ should be larger than
$f_\text{score}\left(x\right)$. Note that it is the ranking of these that are important, and how much larger one is over the other is not that interesting, since the following linear program will find the best one. Another way to motivate this loss is to derive it from a binary classifier trained using a sigmoid activation and cross entropy loss to produce whether $x^*$ or $x$ is the correct solution. Such a detector would be trained using both positive and negative samples. The loss for a positive sample, $\left(x^*, x\right)$, is in that case the same as in Equation~\ref{eq:loss}.
For a negative sample, $\left(x, x^*\right)$, the loss is
\begin{equation}
-\log \left( 1 - \sigma \left(
	    f_\text{score}\left(x\right) - f_\text{score}\left(x^*\right)
	\right) 	\right) ,
\end{equation}
which is also equal to Equation~\ref{eq:loss} since $1-\sigma\left(-x\right) = \sigma\left(x\right)$.

\subsection{Generalized Graph Differences}
\label{sec:ggd}
The feasible solutions, $x^*$ and $x$, can be represented using graphs constructed from the tracking graph by only keeping the edges and vertices with positive flow, i.e. the positive elements of  $x^*$ and $x$.
All terms common to both $x^*$ and $x$ will cancel each other out in the difference in Equation~\ref{eq:loss}. This means that only the terms that differ needs to be considered. That will be refered to as a generalized graph difference. It consists of
\begin{itemize}
\item A set of edges consisting of the edges in $x^*$ but not in $x$ with the same weights and the edges in $x$ but not in $x^*$ with negated weights.
\item A set of vertices consisting of the vertices in $x^*$ but not in $x$ with the same weights and the vertices in $x$ but not in $x^*$ with negated weights.
\end{itemize}
Note that a generalized graph difference is typically not a graph, nor a generalized graph as it can contain edges not connected to any vertices. Thus the focus here is on the graph-difference in a general sense when referring to general graph difference.

This is interesting because hard examples consists of cases where $x^*$ and $x$ are very similar and thus have a lot of terms in common, resulting in small generalized graph differences. Such differences can be constructed by looking at annotated sequences where the optimal solution is known and introducing small errors by changing one or a few edges to form another feasible solution. Table \ref{tab:subgraphs}, \ref{tab:subgraphs2}, \ref{tab:subgraphs3} and \ref{tab:subgraphs4} shows the set of modifications considered and are presented for clarity and reproducibility of our results. Applying these modifications to every position where they apply in a ground truth graph results in a lot of generalized graph differences that can be generated and used for training.

Each generated training example is constructed by taking a single ground truth graph and applying a single modification (from Table 1,2, 3 or 4) to a single position. Hence each batch will contain one mistake per example. All possible such examples are generated and will form one epoch. Note that all these examples can be considered to be hard examples. They are however sufficient to train the system without also introducing easy example or any additional form of bootstrapping or hard mining.

This allows the embedding to rank different kind of tracks. To also allow it to relate a track to the empty solution of there being no track at all, additional generalized graph differences are introduced. Each ground truth track in the graph is split into $n_\text{minlen}$ long non-overlapping subtracks. The first half of each such subtrack is trained to be less than the empty solution and the full subtrack is trained to be greater than the empty solution, see Table~\ref{tab:subgraphs4}.

\begin{table}[hbtp]

\scriptsize

\begin{center}
\centering

\resizebox{0.55\columnwidth}{!}{%
\begin{tabular}{@{}ccc@{}}
\toprule
Name & Ground Truth & Possible Error \\
\midrule
\\
\makecell{ID\\Switch}
&
\begin{tabular}{c}
\begin{tikzpicture}
\node[draw,circle] (A) at (0,0) {$ $};
\node[draw,circle] (B) at (1,0) {$ $};
\node[draw,circle] (C) at (0,1) {$ $};
\node[draw,circle] (D) at (1,1) {$ $};
\draw[->] (A) -> (B);
\draw[->] (C) -> (D);
\end{tikzpicture}
\end{tabular}
&
\begin{tabular}{c}
\begin{tikzpicture}
\node[draw,circle] (A) at (0,0) {$ $};
\node[draw,circle] (B) at (1,0) {$ $};
\node[draw,circle] (C) at (0,1) {$ $};
\node[draw,circle] (D) at (1,1) {$ $};
\draw[->] (A) -> (D);
\draw[->] (C) -> (B);
\end{tikzpicture}
\end{tabular}
\\

\\
Split
&
\begin{tabular}{c}
\begin{tikzpicture}
\node[draw,circle] (A) at (0,0) {$ $};
\node[draw,circle](C) at (1,0) {$ $};
\path[->,draw,thick]
    (A) edge node {$$} (C)  ;
\end{tikzpicture}
\end{tabular}
&
\begin{tabular}{c}
\begin{tikzpicture}
\node[draw,circle] (A) at (0,0) {$ $};
\node (B) at (0.5,0.5) {$ $};
\node[draw,circle](C) at (1,0) {$ $};
\path[->,draw,thick]
    (A) edge [bend right=20]  node {$$} (B)  ;
\path[->,draw,thick]
    (B) edge [bend right=20]  node {$$} (C)  ;
\end{tikzpicture}
\end{tabular}
\\

\\
Merge
&
\begin{tabular}{c}
\begin{tikzpicture}
\node[draw,circle] (A) at (0,0) {$ $};
\node (B) at (0.5,0.5) {$ $};
\node[draw,circle](C) at (1,0) {$ $};
\path[->,draw,thick]
    (A) edge [bend right=20]  node {$$} (B)  ;
\path[->,draw,thick]
    (B) edge [bend right=20]  node {$$} (C)  ;
\end{tikzpicture}
\end{tabular}
&
\begin{tabular}{c}
\begin{tikzpicture}
\node[draw,circle] (A) at (0,0) {$ $};
\node[draw,circle](C) at (1,0) {$ $};
\path[->,draw,thick]
    (A) edge node {$$} (C)  ;
\end{tikzpicture}
\end{tabular}
\\

\\
\makecell{Split\\and\\Merge}
&
\begin{tabular}{c}
\begin{tikzpicture}
\node[draw,circle] (A) at (0,0) {$ $};
\node[draw,circle](B) at (1,0) {$ $};
\node[draw,circle](C) at (1,-1) {$ $};
\node (D) at (0.5,0.5) {$ $};
\path[->,draw,thick]
    (D) edge [bend right=20]  node {$$} (C)  ;
\path[->,draw,thick]
    (A) edge node {$$} (B)  ;
\end{tikzpicture}
\end{tabular}
&
\begin{tabular}{c}
\begin{tikzpicture}
\node[draw,circle] (A) at (0,0) {$ $};
\node[draw,circle](B) at (1,0) {$ $};
\node[draw,circle](C) at (1,-1) {$ $};
\node (D) at (0.5,0.5) {$ $};
\path[->,draw,thick]
    (D) edge [bend right=20]  node {$$} (B)  ;
\path[->,draw,thick]
    (A) edge node {$$} (C)  ;
\end{tikzpicture}
\end{tabular}
\\

\\
\makecell{Double\\Split\\and\\Merge}
&
\begin{tabular}{c}
\begin{tikzpicture}
\node[draw,circle] (A) at (0,0) {$ $};
\node (AA) at (0.5,-0.5) {$ $};
\node[draw,circle] (B) at (1,0) {$ $};
\node[draw,circle] (C) at (0,1) {$ $};
\node[draw,circle] (D) at (1,1) {$ $};
\node (DD) at (0.5,1.5) {$ $};
\draw[->] (A) -> (B);
\draw[->] (C) -> (D);
\end{tikzpicture}
\end{tabular}
&
\begin{tabular}{c}
\begin{tikzpicture}
\node[draw,circle] (A) at (0,0) {$ $};
\node (AA) at (0.5,-0.5) {$ $};
\node[draw,circle] (B) at (1,0) {$ $};
\node[draw,circle] (C) at (0,1) {$ $};
\node[draw,circle] (D) at (1,1) {$ $};
\node (DD) at (0.5,1.5) {$ $};
\draw[->] (C) -> (B);
\path[->,draw,thick]
    (A) edge [bend left=20] node {$$} (AA)  ;
\path[->,draw,thick]
    (DD) edge [bend right=20] node {$$} (D)  ;
\end{tikzpicture}
\end{tabular}
\\
\bottomrule
\end{tabular}
}
\end{center}
\normalsize
\caption{Switch, split and merge errors introduced to form training data pairs from ground truth.}
\label{tab:subgraphs}

\end{table}

\begin{table}
\scriptsize
\begin{center}
\centering
\resizebox{0.75\columnwidth}{!}{%
\begin{tabular}{@{}ccc@{}}
\toprule
 Name & Ground Truth & Possible Error \\
\midrule

\\
\makecell{Detection\\Skip}
&
\begin{tabular}{c}
\begin{tikzpicture}
\node[draw,circle] (A) at (0,0) {$ $};
\node[draw,circle] (B) at (1,0) {$ $};
\node[draw,circle] (C) at (2,0) {$ $};
\path[->,draw,thick]
    (A) edge node {$$} (B)  ;
\path[->,draw,thick]
    (B) edge node {$$} (C)  ;
\end{tikzpicture}
\end{tabular}
&
\begin{tabular}{c}
\begin{tikzpicture}
\node[draw,circle] (A) at (0,0) {$ $};
\node[draw,circle] (B) at (1,0) {$ $};
\node[draw,circle] (C) at (2,0) {$ $};
\path[->,draw,thick]
    (A) edge [bend left=30]  node {$$} (C)  ;
\end{tikzpicture}
\end{tabular}
\\

\\
\makecell{Skip\\First}
&
\begin{tabular}{c}
\begin{tikzpicture}
\node[draw,circle] (A) at (0,0) {$ $};
\node (B) at (-0.5,0.5) {$ $};
\node[draw,circle] (C) at (1,0) {$ $};
\path[->,draw,thick]
    (B) edge [bend right=20] node {$$} (A)  ;
\draw[->] (A) -> (C);
\end{tikzpicture}
\end{tabular}
&
\begin{tabular}{c}
\begin{tikzpicture}
\node[draw,circle] (A) at (0,0) {$ $};
\node (B) at (0.5,0.5) {$ $};
\node[draw,circle] (C) at (1,0) {$ $};
\path[->,draw,thick]
    (B) edge [bend right=20] node {$$} (C)  ;
\end{tikzpicture}
\end{tabular}
\\

\\
\makecell{Skip\\Last}
&
\begin{tabular}{c}
\begin{tikzpicture}
\node[draw,circle] (A) at (0,0) {$ $};
\node (B) at (1.5,-0.5) {$ $};
\node (D) at (1.5,0.5) {$ $};
\node[draw,circle] (C) at (1,0) {$ $};
\path[->,draw,thick]
    (C) edge [bend left=20] node {$$} (B)  ;
\draw[->] (A) -> (C);
\end{tikzpicture}
\end{tabular}
&
\begin{tabular}{c}
\begin{tikzpicture}
\node[draw,circle] (A) at (0,0) {$ $};
\node (B) at (0.5,0.5) {$ $};
\node (D) at (0.5,-0.5) {$ $};
\node[draw,circle] (C) at (1,0) {$ $};
\path[->,draw,thick]
    (A) edge [bend left=20] node {$$} (D)  ;
\end{tikzpicture}
\end{tabular}
\\

\\
\makecell{Extra\\First}
&
\begin{tabular}{c}
\begin{tikzpicture}
\node[draw,circle] (A) at (0,0) {$ $};
\node (B) at (0.5,-0.5) {$ $};
\node (D) at (0.5,0.5) {$ $};
\node[draw,circle] (C) at (1,0) {$ $};
\path[->,draw,thick]
    (D) edge [bend right=20] node {$$} (C)  ;
\end{tikzpicture}
\end{tabular}
&
\begin{tabular}{c}
\begin{tikzpicture}
\node[draw,circle] (A) at (0,0) {$ $};
\node (B) at (-0.5,0.5) {$ $};
\node (D) at (-0.5,-0.5) {$ $};
\node[draw,circle] (C) at (1,0) {$ $};
\path[->,draw,thick]
    (B) edge [bend right=20] node {$$} (A)  ;
\draw[->] (A) -> (C);
\end{tikzpicture}
\end{tabular}
\\

\\
\makecell{Extra\\Last}
&
\begin{tabular}{c}
\begin{tikzpicture}
\node[draw,circle] (A) at (0,0) {$ $};
\node[draw,circle] (B) at (1,0) {$ $};
\node (C) at (0.5,-0.5) {$ $};
\path[->,draw,thick]
    (A) edge [bend right=20]  node {$$} (C)  ;
\end{tikzpicture}
\end{tabular}
&
\begin{tabular}{c}
\begin{tikzpicture}
\node[draw,circle] (A) at (0,0) {$ $};
\node[draw,circle] (B) at (1,0) {$ $};
\node (C) at (1.5,-0.5) {$ $};
\path[->,draw,thick]
    (A) edge  node {$$} (B)  ;
\path[->,draw,thick]
    (B) edge [bend right=20] node {$$} (C)  ;
\end{tikzpicture}
\end{tabular}
\\
\bottomrule

\end{tabular}
}
\end{center}
\normalsize
\caption{Skip and extra errors introduced to form training data pairs from ground truth.}
\label{tab:subgraphs2}
\end{table}

\begin{table}
\scriptsize
\begin{center}
\centering
\resizebox{0.75\columnwidth}{!}{%
\begin{tabular}{@{}ccc@{}}
\toprule
Name & Ground Truth & Possible Error \\
\midrule
\\
\makecell{False\\positive}
&
\begin{tabular}{c}
\begin{tikzpicture}
\end{tikzpicture}
\end{tabular}
&
\begin{tabular}{c}
\begin{tikzpicture}
\node (A) at (0,0) {$ $};
\node[draw,circle,dashed] (B) at (0.5,-0.5) {$ $};
\node(C) at (1,0) {$ $};
\path[->,draw,thick]
    (A) edge [bend right=20]  node {$$} (B)  ;
\path[->,draw,thick]
    (B) edge [bend right=20]  node {$$} (C)  ;
\end{tikzpicture}
\end{tabular}
\\

\\
\makecell{Split\\to\\False\\ Positive}
&
\begin{tabular}{c}
\begin{tikzpicture}
\node[draw,circle] (A) at (0,0) {$ $};
\node[draw,circle] (B) at (1,0) {$ $};
\node[draw,circle] (C) at (2,0) {$ $};
\node[draw,circle,dashed] (D) at (2,-1) {$ $};
\node (E) at (1.5,0.5) {$ $};
\node (F) at (2.5,-1.5) {$ $};
\path[->,draw,thick]
    (A) edge  node {$$} (B)  ;
\path[->,draw,thick]
    (B) edge  node {$$} (C)  ;
\end{tikzpicture}
\end{tabular}
&
\begin{tabular}{c}
\begin{tikzpicture}
\node[draw,circle] (A) at (0,0) {$ $};
\node[draw,circle] (B) at (1,0) {$ $};
\node[draw,circle] (C) at (2,0) {$ $};
\node[draw,circle,dashed] (D) at (2,-1) {$ $};
\node (E) at (1.5,0.5) {$ $};
\node (F) at (2.5,-1.5) {$ $};
\path[->,draw,thick]
    (A) edge  node {$$} (B)  ;
\path[->,draw,thick]
    (B) edge  node {$$} (D)  ;
\path[->,draw,thick]
    (E) edge [bend right=20] node {$$} (C)  ;
\path[->,draw,thick]
    (D) edge [bend right=20] node {$$} (F)  ;
\end{tikzpicture}
\end{tabular}
\\

\\
\makecell{Split\\from\\False\\ Positive}
&
\begin{tabular}{c}
\begin{tikzpicture}
\node[draw,circle] (A) at (0,0) {$ $};
\node[draw,circle] (B) at (1,0) {$ $};
\node[draw,circle] (C) at (2,0) {$ $};
\node[draw,circle,dashed] (D) at (1,-1) {$ $};
\node (E) at (1.5,0.5) {$ $};
\node (F) at (2.5,-1.5) {$ $};
\path[->,draw,thick]
    (A) edge  node {$$} (B)  ;
\path[->,draw,thick]
    (B) edge  node {$$} (C)  ;
\end{tikzpicture}
\end{tabular}
&
\begin{tabular}{c}
\begin{tikzpicture}
\node[draw,circle] (A) at (0,0) {$ $};
\node[draw,circle] (B) at (1,0) {$ $};
\node[draw,circle] (C) at (2,0) {$ $};
\node[draw,circle,dashed] (D) at (1,-1) {$ $};
\node (E) at (1.5,0.5) {$ $};
\node (F) at (0.5,-1.5) {$ $};
\path[->,draw,thick]
    (A) edge  node {$$} (B)  ;
\path[->,draw,thick]
    (B)  edge [bend right=20]  node {$$} (E)  ;
\path[->,draw,thick]
    (F) edge [bend left=20] node {$$} (D)  ;
\path[->,draw,thick]
    (D) edge node {$$} (C)  ;
\end{tikzpicture}
\end{tabular}
\\
\bottomrule

\end{tabular}
}
\end{center}
\normalsize
\caption{False positives introduced to form training data pairs from ground truth.}
\label{tab:subgraphs3}
\end{table}

\begin{table}
\scriptsize
\begin{center}
\centering
\resizebox{0.75\columnwidth}{!}{%
\begin{tabular}{ccc}
\toprule
Name & Ground Truth & Possible Error \\
\midrule

\\
\makecell{Too Short\\Track}
&
\begin{tabular}{c}
\begin{tikzpicture}
\end{tikzpicture}
\end{tabular}
&
\begin{tabular}{c}
\begin{tikzpicture}[scale=0.75]
\node[draw,circle] (A) at (0,0) {$ $};
\node[draw,circle] (B) at (1,0) {$ $};
\node[draw,circle] (C) at (2,0) {$ $};
\node (D) at (1.5,0) {$\ldots$};
\node (E) at (-0.5,0.5) {$ $};
\node (F) at (2.5,0.5) {$ $};
\path[->,draw,thick]
    (A) edge  node {$$} (B)  ;

\path[->,draw,thick]
    (E) edge [bend right=20] node {$$} (A)  ;
\path[->,draw,thick]
    (C) edge [bend right=20] node {$$} (F)  ;

\draw [decorate,decoration={brace,mirror,amplitude=10pt}]
(-0.1,-0.2) -- (2.1,-0.2) node [black,midway,yshift=-17pt] {\footnotesize
$\frac{n_{\text{minlen}}}{2}$};
\end{tikzpicture}
\end{tabular}
\\

\\
\makecell{Proper\\Track}
&
\begin{tabular}{c}
\begin{tikzpicture}[scale=0.75]
\node[draw,circle] (A) at (0,0) {$ $};
\node[draw,circle] (B) at (1,0) {$ $};
\node[draw,circle] (C) at (2,0) {$ $};
\node (D) at (1.5,0) {$\ldots$};
\node (E) at (-0.5,0.5) {$ $};
\node (F) at (2.5,0.5) {$ $};
\path[->,draw,thick]
    (A) edge  node {$$} (B)  ;

\path[->,draw,thick]
    (E) edge [bend right=20] node {$$} (A)  ;
\path[->,draw,thick]
    (C) edge [bend right=20] node {$$} (F)  ;

\draw [decorate,decoration={brace,mirror,amplitude=10pt}]
(-0.1,-0.2) -- (2.1,-0.2) node [black,midway,yshift=-17pt] {\footnotesize
$n_{\text{minlen}}$};
\end{tikzpicture}
\end{tabular}
&
\begin{tabular}{c}
\begin{tikzpicture}
\end{tikzpicture}
\end{tabular}
\\
\bottomrule
\end{tabular}
}
\end{center}
\normalsize
\caption{Track lengths data pairs used for training.}
\label{tab:subgraphs4}
\end{table}


\section{Experiments}
\label{sec:experiments}

In this section, performance of the proposed methods is presented. The code used for the evaluation is made availible\footnote{\url{https://github.com/hakanardo/ggdtrack}}.

\subsection{Generalized graph difference training}

\begin{table}
\centering
\ra{1.3}
\begin{tabular}{@{}lrrrr@{}}
\toprule
& MOTA & MT & IDS  &  FRAG  \\
\midrule
Linear [23] & 28.25 & 9.67 & 342 & 1620 \\
MLP1 [23] & 31.05 & 8.32 & {\bf 282} & 1553\\
MLP2 [23] & {\bf 31.10} & {\bf 8.51} & 289 & 1562\\
\midrule
Linear GGD & 27.08 & 5.99 & 401 & {\bf 539} \\
MLP1 GGD & 30.77 & {\bf 8.51} & 442 & 805 \\
MLP2 GGD & 31.06 & {\bf 8.51} & 423 & 802 \\
\bottomrule
\end{tabular}
\caption{Comparing the proposed traning scheme (GGD) with
[23] using Multiple Object Tracking Accuracy (MOTA), Mostly Tracked (MT), ID Switches (IDS) and Fragmentations (FRAG).}
\label{tab:opt}
\end{table}

\begin{table*}
\centering
\ra{1.3}
\begin{tabular}{@{}lrrrrrcrrrrr@{}}
\toprule
& \multicolumn{5}{c}{Easy} & \phantom{abc}& \multicolumn{5}{c}{Hard} \\
\cmidrule{2-6} \cmidrule{8-12}
& MOTA & MOTP & IDF1  &  IDP  &  IDR & & MOTA & MOTP & IDF1  &  IDP  &  IDR  \\
\midrule

MYTRACKER \cite{8387029}&
{\bf 78.3} & 78.4 & {\bf 80.3} & 87.3 & {\bf 74.4} & &
59.6 & 76.7 & 63.5 & 73.9 & 55.6	 \\

Proposed  &
74.0 & 75.2 & 71.8 & 79.1 & 65.7 & &
{\bf 63.1} & 74.4 & 63.3 & 73.7 & 55.4 \\


MTMC CDSC \cite{DBLP:journals/corr/TesfayeZPPS17} &
70.9 & 75.8	& 77.0 & 87.6	& 68.6	& &
59.6	& 75.4	& 65.5	& 81.4	& 54.7	\\

MTMC ReIDp \cite{DBLP:journals/corr/abs-1712-09531} &
68.8 & 77.9	& 79.2 & {\bf 89.9} & 70.7 & &
60.9 & 76.8	& {\bf 71.6} & {\bf 85.3} & {\bf 61.7} \\

BIPCC \cite{DBLP:conf/eccv/RistaniSZCT16} &
59.4 & {\bf 78.7} & 70.1 & 83.6 & 60.4 & &
54.6 & {\bf 77.1} & 64.5 & 81.2 & 53.5	\\

PT BIPCC \cite{8237540} &
59.3 & {\bf 78.7} & 71.2 & 84.8 & 61.4 & &
54.4 & {\bf 77.1} & 65.0 & 81.8 & 54.0 \\

\bottomrule
\end{tabular}
\caption{Results on DukeMTMC single camera test set.}
\label{tab:dukeresult}
\end{table*}

Initialy, the main difference between integrating the linear programming solver into the training process versus the proposed ranking loss will be investigated. 

To facilitate this an experiment performed by Schulter et al. \cite{Schulter2017DeepNF} will be recosntructed. There vertex weights are formed by a linear network taking the detecion score as input and three different networks (linear, one layer MLP with 64 hidden features and two layer MLP with 32 hidden features in each layer) are tried to produce the edge weights from features consisting of bounding boxe differences, detection confidences, temporal differences, and the IoU values. Results are evaluated on MOTA16 and presented in Tab~\ref{tab:mota16} and show
that the two approaches perform similarly. This shows that the simpler approach of generating generalized graph differences is capable of learning at least as much from the traning data as the more advanced approach of integrating the linear program into the training process.

\begin{figure}
\centering
\includegraphics[width=0.8\columnwidth]{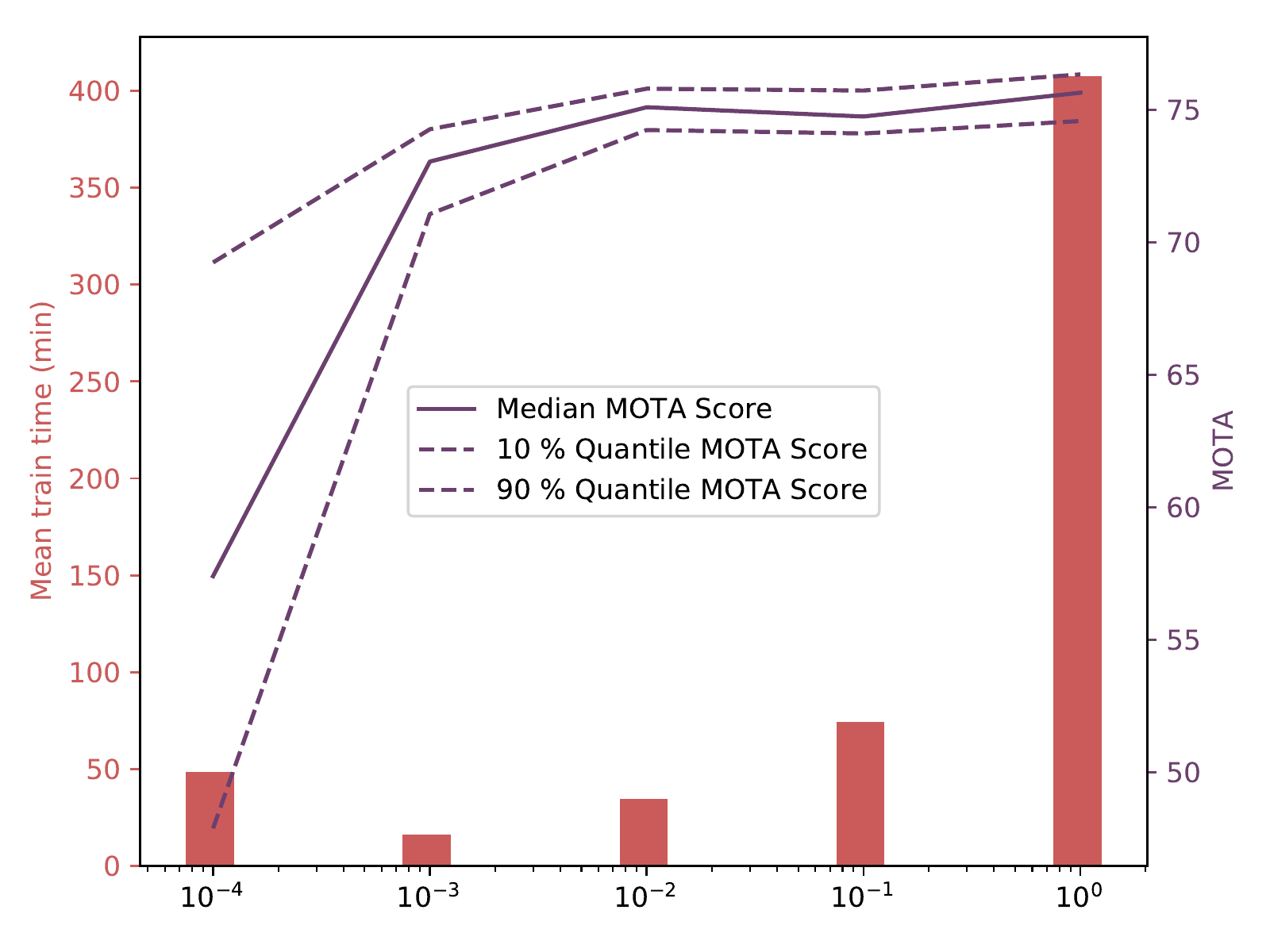}
\caption{MOTA score (median and quantiles over ten runs) of the proposed tracker trained on different amounts of training data (purple) and execution time for the training (red).}
\label{fig:learning_curve}
\end{figure}

\subsection{DukeMTMC}
Evaluation is performed, similar to other works, on the DukeMTMC \cite{Ristani2014,DBLP:conf/eccv/RistaniSZCT16} single camera tracking challenge.
It consists of $60$ fps recordings from $8$ different cameras showing pedestrians moving around a campus.
Public detections are provided by DukeMTMCT. They were produced using Felzenszwalb et al.'s Deformable Part Models (DPM) \cite{DBLP:conf/cvpr/FelzenszwalbMR08} and are used in this evaluation.

Some of the meta parameters use were made to depend on the framerate, $n_\text{fps}=60$. The values used were $r_\text{neighbours}=5$, $t_\text{max}=3n_\text{fps}$, $n_\text{velest}=0.5n_\text{fps}$, $n_\text{project}=n_\text{fps}$ and $n_\text{minlen}=2$. The model architecture parameters were set to $n_\text{detlayers}=4$, $n_\text{detfeat}=32$, $n_\text{kltlayers}=7$, $n_\text{kltfeat}=64$, $n_\text{longlayers}=7$, $n_\text{longfeat}=32$, $n_\text{combinelayers}=4$, $n_\text{combinefeat}=256$.

The dataset contains both an easy and a hard test set, where the hard test set shows larger crowds moving around as opposed to the more normal pedestrian flows in the easy test set.
The standard metrics and protocol for evaluation is utilized \cite{Bernardin2008,Li2009,DBLP:conf/eccv/RistaniSZCT16}.

Results are presented in Table~\ref{tab:dukeresult}, where the proposed method is compared to other state of the art methods also based on the public DPM detections of the dataset.

These results were achieved by training on the first $90$\% of the trainval part of the dataset and using the remaining $10$\% as a validation set. Training were performed using the Adam optimizer \cite{DBLP:journals/corr/KingmaB14} for ten epochs. The videos were processed in chunks of $600$ frames each, with an overlap of $60$ frames, forming a single tracking graph for each such chunk. The overlap was used to match ID-numbers between the chunks of the test data by solving the assignment problem with a cost matrix containing negated counts of how many  detections the tracks had in common. The tracks produced are sequences of detections found by the detector and will thus contain holes where the detector have failed. These holes are filled by linear interpolation.

The MOTA score of the proposed algorithm is very competitive.
It outperforms the current state of the art on the hard test set while being second best on the easy test set.
At the same time, the IDF1 score is trailing behind. This is to be expected with this kind of tracker that does not contain any form of appearance modelling to do re-identification to correct ID-switch mistakes.

\subsection{Impact of Training Dataset Size}
In order to show how efficient the training data is utilized by the proposed generalized graph diff data augmentation, training were also performed on reduced training data sets. Five cases were investigated by randomly sampling $100\%$, $10\%$, $1\%$, $0.1\%$ and $0.01\%$ of the generated generalized graph differences constructed form the training data (the first $90\%$ of trainval).
Each sampling was repeated $10$ times and different models was trained from the different subsets for each case.  All models were then evaluated on the full validation set (the last $10\%$ of trainval). Training was allowed to continue until the validation accuracy on general graph differences did not improve for three consecutive epochs, and the best performing epoch was saved. This allowed models trained on a small amount of data to use more epochs if needed.
The median MOTA scores for each case and their quantiles are plotted in Figure~\ref{fig:learning_curve}. The IDF1 scores show the same carecteristics.

With only $1\%$ of the training data, the MOTA score is almost identical to the case utilizing all the data (75.1 instead of 75.65), and the training time is reduced over 10 times to only 34 minutes on average on a 2.20GHz Xeon CPU E5-2630v4 using a single GTX 1080 GPU. Note that this is the training time only and does not include the preprocessing producing the general graph differences.

%








\section{Conclusions}

We have presented a method that can learn the weights of a network flow tracker from generalized graph differences. That is an efficient representation of differences between graphs. Training data were produced from small perturbation of ground truth tracks which allows the model to be trained using the standard Adam optimizer. There is no need to solve an additional optimisation problem for each example in each training batch as prior work do. The method was evaluated on the challenging DukeMTMCT dataset and showed very competitive results in the MOTA metric, especially on the hard test set where it outperformed the state of the art. Also it is capable of learning a competitive model, with only $0.55$ lower MOTA score, using only $1\%$ of the training data.


{\small
\bibliographystyle{ieee}
\bibliography{refs}

\begin{thebibliography}{10}\itemsep=-1pt

\bibitem{Ahuja:1993:NFT:137406}
R.~K. Ahuja, T.~L. Magnanti, and J.~B. Orlin.
\newblock {\em Network Flows: Theory, Algorithms, and Applications}.
\newblock Prentice-Hall, Inc., Upper Saddle River, NJ, USA, 1993.

\bibitem{Berclaz2011}
J.~Berclaz, F.~Fleuret, E.~Turetken, and P.~Fua.
\newblock Multiple object tracking using k-shortest paths optimization.
\newblock {\em IEEE Transactions on Pattern Analysis and Machine Intelligence},
  33(9):1806--1819, Sept 2011.

\bibitem{Bernardin2008}
K.~Bernardin and R.~Stiefelhagen.
\newblock Evaluating multiple object tracking performance: The clear mot
  metrics.
\newblock {\em EURASIP Journal on Image and Video Processing}, 2008(1):246309,
  May 2008.

\bibitem{Bertsekas:1992:DN:121104}
D.~Bertsekas and R.~Gallager.
\newblock {\em Data Networks (2Nd Ed.)}.
\newblock Prentice-Hall, Inc., Upper Saddle River, NJ, USA, 1992.

\bibitem{Cao2017}
Z.~Cao, T.~Simon, S.~Wei, and Y.~Sheikh.
\newblock Realtime multi-person 2d pose estimation using part affinity fields.
\newblock In {\em 2017 IEEE Conference on Computer Vision and Pattern
  Recognition (CVPR)}, pages 1302--1310, July 2017.

\bibitem{Jifeng2016}
J.~Dai, Y.~Li, K.~He, and J.~Sun.
\newblock R-fcn: Object detection via region-based fully convolutional
  networks.
\newblock In D.~D. Lee, M.~Sugiyama, U.~V. Luxburg, I.~Guyon, and R.~Garnett,
  editors, {\em Advances in Neural Information Processing Systems 29}, pages
  379--387. Curran Associates, Inc., 2016.

\bibitem{DBLP:conf/cvpr/FelzenszwalbMR08}
P.~F. Felzenszwalb, D.~A. McAllester, and D.~Ramanan.
\newblock A discriminatively trained, multiscale, deformable part model.
\newblock In {\em 2008 {IEEE} Computer Society Conference on Computer Vision
  and Pattern Recognition {(CVPR} 2008), 24-26 June 2008, Anchorage, Alaska,
  {USA}}, 2008.

\bibitem{Frossard2018}
D.~Frossard and R.~Urtasun.
\newblock End-to-end learning of multi-sensor 3d tracking by detection.
\newblock In {\em 2018 IEEE International Conference on Robotics and Automation
  (ICRA)}, pages 635--642, May 2018.

\bibitem{Kim2015}
C.~Kim, F.~Li, A.~Ciptadi, and J.~M. Rehg.
\newblock Multiple hypothesis tracking revisited.
\newblock In {\em 2015 IEEE International Conference on Computer Vision
  (ICCV)}, pages 4696--4704, Dec 2015.

\bibitem{DBLP:journals/corr/KingmaB14}
D.~P. Kingma and J.~Ba.
\newblock Adam: {A} method for stochastic optimization.
\newblock {\em CoRR}, abs/1412.6980, 2014.

\bibitem{Li2009}
Y.~Li, C.~Huang, and R.~Nevatia.
\newblock Learning to associate: Hybridboosted multi-target tracker for crowded
  scene.
\newblock In {\em 2009 IEEE Conference on Computer Vision and Pattern
  Recognition}, pages 2953--2960, June 2009.

\bibitem{Lin2017}
T.~Lin, P.~Dollár, R.~Girshick, K.~He, B.~Hariharan, and S.~Belongie.
\newblock Feature pyramid networks for object detection.
\newblock In {\em 2017 IEEE Conference on Computer Vision and Pattern
  Recognition (CVPR)}, pages 936--944, July 2017.

\bibitem{Liu2016}
W.~Liu, D.~Anguelov, D.~Erhan, C.~Szegedy, S.~Reed, C.~Fu, and A.~Berg.
\newblock Ssd: Single shot multibox detector.
\newblock In B.~Leibe, J.~Matas, M.~Welling, and N.~Sebe, editors, {\em
  Computer Vision - 14th European Conference, ECCV 2016, Proceedings}, Lecture
  Notes in Computer Science (including subseries Lecture Notes in Artificial
  Intelligence and Lecture Notes in Bioinformatics), pages 21--37, Germany, 1
  2016. Springer Verlag.

\bibitem{Luo14}
W.~Luo, X.~Zhao, and T.~Kim.
\newblock Multiple object tracking: {A} review.
\newblock {\em CoRR}, abs/1409.7618, 2014.

\bibitem{1261119}
R.~P.~S. Mahler.
\newblock Multitarget bayes filtering via first-order multitarget moments.
\newblock {\em IEEE Transactions on Aerospace and Electronic Systems},
  39(4):1152--1178, Oct 2003.

\bibitem{Mahler:2007:SMI:1512927}
R.~P.~S. Mahler.
\newblock {\em Statistical Multisource-Multitarget Information Fusion}.
\newblock Artech House, Inc., Norwood, MA, USA, 2007.

\bibitem{8237540}
A.~Maksai, X.~Wang, F.~Fleuret, and P.~Fua.
\newblock Non-markovian globally consistent multi-object tracking.
\newblock In {\em 2017 IEEE International Conference on Computer Vision
  (ICCV)}, pages 2563--2573, Oct 2017.

\bibitem{Pirsiavash:2011:GGA:2191740.2191761}
H.~Pirsiavash, D.~Ramanan, and C.~C. Fowlkes.
\newblock Globally-optimal greedy algorithms for tracking a variable number of
  objects.
\newblock In {\em Proceedings of the 2011 IEEE Conference on Computer Vision
  and Pattern Recognition}, CVPR '11, pages 1201--1208, Washington, DC, USA,
  2011. IEEE Computer Society.

\bibitem{Redmon2017}
J.~Redmon and A.~Farhadi.
\newblock Yolo9000: Better, faster, stronger.
\newblock In {\em 2017 IEEE Conference on Computer Vision and Pattern
  Recognition (CVPR)}, pages 6517--6525, July 2017.

\bibitem{DBLP:conf/eccv/RistaniSZCT16}
E.~Ristani, F.~Solera, R.~S. Zou, R.~Cucchiara, and C.~Tomasi.
\newblock Performance measures and a data set for multi-target, multi-camera
  tracking.
\newblock In {\em Computer Vision - {ECCV} 2016 Workshops - Amsterdam, The
  Netherlands, October 8-10 and 15-16, 2016, Proceedings, Part {II}}, pages
  17--35, 2016.

\bibitem{Ristani2014}
E.~Ristani and C.~Tomasi.
\newblock Tracking multiple people online and in real time.
\newblock In {\em Asian Conference on Computer Vision}, pages 444--459.
  Springer, 2014.

\bibitem{Schulter2017DeepNF}
S.~Schulter, P.~Vernaza, W.~Choi, and M.~K. Chandraker.
\newblock Deep network flow for multi-object tracking.
\newblock {\em 2017 IEEE Conference on Computer Vision and Pattern Recognition
  (CVPR)}, pages 2730--2739, 2017.

\bibitem{DBLP:journals/corr/TesfayeZPPS17}
Y.~T. Tesfaye, E.~Zemene, A.~Prati, M.~Pelillo, and M.~Shah.
\newblock Multi-target tracking in multiple non-overlapping cameras using
  constrained dominant sets.
\newblock {\em CoRR}, abs/1706.06196, 2017.

\bibitem{xiao2018simple}
B.~Xiao, H.~Wu, and Y.~Wei.
\newblock Simple baselines for human pose estimation and tracking.
\newblock In {\em European Conference on Computer Vision (ECCV)}, 2018.

\bibitem{8387029}
K.~Yoon, Y.~Song, and M.~Jeon.
\newblock Multiple hypothesis tracking algorithm for multi-target multi-camera
  tracking with disjoint views.
\newblock {\em IET Image Processing}, 12(7):1175--1184, 2018.

\bibitem{Bouguet00pyramidalimplementation}
J.~yves Bouguet.
\newblock Pyramidal implementation of the lucas kanade feature tracker.
\newblock {\em Intel Corporation, Microprocessor Research Labs}, 2000.

\bibitem{DBLP:conf/cvpr/ZhangLN08}
L.~Zhang, Y.~Li, and R.~Nevatia.
\newblock Global data association for multi-object tracking using network
  flows.
\newblock In {\em 2008 {IEEE} Computer Society Conference on Computer Vision
  and Pattern Recognition {(CVPR} 2008), 24-26 June 2008, Anchorage, Alaska,
  {USA}}, 2008.

\bibitem{DBLP:journals/corr/abs-1712-09531}
Z.~Zhang, J.~Wu, X.~Zhang, and C.~Zhang.
\newblock Multi-target, multi-camera tracking by hierarchical clustering:
  Recent progress on dukemtmc project.
\newblock {\em CoRR}, abs/1712.09531, 2017.

\end{thebibliography}
}

\end{document}